\documentclass[preprint,review,3p]{elsarticle}

\usepackage{amssymb}
\usepackage{lineno}
\input{macros.tex}
\journal{Arxiv}

\begin{document}

% \linenumbers
\begin{frontmatter}

    %% Title, authors and addresses
    
    %% use the tnoteref command within \title for footnotes;
    %% use the tnotetext command for theassociated footnote;
    %% use the fnref command within \author or \address for footnotes;
    %% use the fntext command for theassociated footnote;
    %% use the corref command within \author for corresponding author footnotes;
    %% use the cortext command for theassociated footnote;
    %% use the ead command for the email address,
    %% and the form \ead[url] for the home page:
    %% \title{Title\tnoteref{label1}}
    %% \tnotetext[label1]{}
    %% \author{Name\corref{cor1}\fnref{label2}}
    %% \ead{email address}
    %% \ead[url]{home page}
    %% \fntext[label2]{}
    %% \cortext[cor1]{}
    %% \affiliation{organization={},
    %%             addressline={},
    %%             city={},
    %%             postcode={},
    %%             state={},
    %%             country={}}
    %% \fntext[label3]{}
    
    \title{Cost-Effective Retraining of Machine Learning Models}
    
    %% use optional labels to link authors explicitly to addresses:
    %% \author[label1,label2]{}
    %% \affiliation[label1]{organization={},
    %%             addressline={},
    %%             city={},
    %%             postcode={},
    %%             state={},
    %%             country={}}
    %%
    %% \affiliation[label2]{organization={},
    %%             addressline={},
    %%             city={},
    %%             postcode={},
    %%             state={},
    %%             country={}}
    
    \author{Ananth Mahadevan}
    \author{Michael Mathioudakis}
    \affiliation{organization={Department of Computer Science, University of Helsinki},%Department and Organization
                % addressline={}, 
                % city={Helsinki},
                % postcode={}, 
                % state={},
                country={Finland}}

    %% Text of abstract
    \begin{abstract}
    It is important to retrain a machine learning (ML) model in order to maintain its performance as the data changes over time.
    However, this can be costly as it usually requires processing the entire dataset again.
    This creates a trade-off between retraining too frequently, which leads to unnecessary computing costs, and not retraining often enough, which results in stale and inaccurate ML models. 
    To address this challenge, we propose ML systems that make automated and cost-effective decisions about when to retrain an ML model.
    We aim to optimize the trade-off by considering the costs associated with each decision.
    Our research focuses on determining whether to retrain or keep an existing ML model based on various factors, including the data, the model, and the predictive queries answered by the model. 
    Our main contribution is a Cost-Aware Retraining Algorithm called \cara, which optimizes the trade-off over streams of data and queries.
    To evaluate the performance of \cara, we analyzed synthetic datasets and demonstrated that 
    \cara can adapt to different data drifts and retraining costs while performing similarly to an optimal retrospective algorithm.
    We also conducted experiments with real-world datasets and showed that Cara achieves better accuracy than drift detection baselines while making fewer retraining decisions, ultimately resulting in lower total costs.
\end{abstract}
    
    %%Graphical abstract
    % \begin{graphicalabstract}
    % %\includegraphics{grabs}
    % \end{graphicalabstract}
    
    %%Research highlights
    % \begin{highlights}
    % \item Research highlight 1
    % \item Research highlight 2
    % \end{highlights}
    
    \begin{keyword}
        machine learning \sep resource-aware computing
    %% keywords here, in the form: keyword \sep keyword
    
    %% PACS codes here, in the form: \PACS code \sep code
    
    %% MSC codes here, in the form: \MSC code \sep code
    %% or \MSC[2008] code \sep code (2000 is the default)
    
    \end{keyword}
    
    \end{frontmatter}
\section{Introduction}
\label{sec:intro}

Retraining a machine learning (ML) model is essential in the presence of \emph{data drift} \cite{gama2014SurveyConceptDrift}, i.e., continuous changes in the data due to factors such as system modifications, seasonality, or changes in user preferences.
As a data drift occurs, the performance of an ML model trained on old data typically decreases -- or, more generally, is not as high as it could be if it took advantage of the new data.
As a result, maintaining the performance of an ML model calls for an online decision on whether to \retrain~or \keep~the existing ML model.
In what follows, we refer to any algorithm that makes such a decision as a \emph{retraining algorithm}.

\revision{
Several methods exist to detect data drift \citep{gama2004LearningDriftDetectiona,bifet2007LearningTimeChangingData,RAAB2020ReactiveSoftKS} by monitoring the errors of an ML model on a stream of labeled samples, i.e., \emph{data}.
These methods decide to retrain the ML model whenever a drift is detected.
%
% research gap
However, note that (i) ML models are typically used to predict a stream of unlabeled samples \citep{chandra2016AdaptiveFrameworkMultistream}, i.e., \emph{queries}, and (ii) (re)training is associated with \emph{costs}, such as the monetary cost for (re)training a model on a cloud service or the energy cost for executing the training algorithm \citep{strubell2020EnergyPolicyConsiderations}.
Considering queries and costs in addition to data is essential in deciding whether to retrain a model.}
To see why, \revision{consider a real estate firm that uses an ML model to predict housing prices. 
Suppose, the data used to train the ML model represents the price distribution of all houses in the market.
However, the customer queries concern only a niche of this market, for example, mansions.
In such a case, if new data suggests prices have changed for studio apartments but not for mansions, retraining the model will not improve its performance on queries significantly enough to justify the retraining cost. 
By contrast, if the new data indicates that prices for mansions have changed significantly, then the potential drop in performance on queries will warrant a \retrain decision.}
%% Old example   
% To see why, consider the scenario where a model is trained to predict the prices of different products in an online marketplace.
%
% Suppose also that most queries request predictions for the prices of textbooks.
%
% In such case, even if new data suggest that prices may have changed for another category, say Rubik's cubes, but not for textbooks, then retraining the model will not improve its overall performance significantly enough to justify the retraining cost.
%
% By contrast, if new data suggest that textbook prices have changed significantly, then the potential drop in performance for textbooks will warrant a \retrain decision.
% 
Unfortunately, many existing \revision{drift detection methods} do not consider queries or retraining costs, which may lead to suboptimal decisions for scenarios like the above.

The above discussion \revision{gives rise to two costs} when deciding whether to retrain.
The first is a \textit{model staleness cost}, i.e., \revision{the performance loss} due to keeping an ML model trained on old data.
Typically, when a data drift occurs, a model trained on old data will have lower performance compared to an ML model trained on fresher data.
Note that while we use the notion of ``cost'' in a general rather than strictly monetary sense, a \revision{performance loss} often does translate into a monetary cost.
For example, in the scenario of a real estate firm described earlier, \revision{a performance loss} would lead to a loss in customer satisfaction and, eventually, a monetary cost.

The second type of cost is the \textit{model retraining cost}, i.e., the resources spent to train an ML model on new data.
Again, this cost may be monetary (e.g., renting a machine from an online cloud service to execute the training) or of other kinds (e.g., energy consumption).
This cost depends on the class of ML models used, the type of data, and the training algorithm.

There is a direct trade-off between these two costs of staleness and retraining.
While retraining frequently results in a fresher ML model, the performance increase may be marginal when compared to the cost of retraining.
On the other hand, \revision{infrequent retraining reduces the retraining cost but results in a stale model with performance loss.}

\spara{Our contributions}
In this paper, we formalize and study the trade-off between staleness and retraining costs.
Our main contribution is an online Cost-Aware Retraining Algorithm (\cara) that optimizes the trade-off between the two costs.
\cara is defined as a function that makes \retrain or \keep decisions based on a choice of retraining parameters that consider both data and queries, along with the costs associated with each decision. 
We present three variants of \cara for different such parameters.
In addition, we present a retrospective optimal algorithm \oracle, which we use as a baseline in our experimental comparisons to provide an upper bound on the performance of online retraining algorithms.
In more detail, in synthetic experiments, we vary both data and query distribution and showcase how \cara captures and adapts to different data drifts, achieving similar performance to an optimal algorithm.
Moreover, using real-world datasets, we compare with standard drift detection baselines from the literature and demonstrate that \cara exhibits better accuracy, but with fewer \retrain decisions, and thus lower total costs. 

\section{Related Work}
\label{sec:related-works}

We identify three main categories of existing work in the literature that are related to our contributions.

\spara{Drift Detection Methods} 
These methods make a \retrain decision based on a signal from a drift detector.
The simplest detectors monitor the model's error in a window and use statistical principles such as Page-Hinkely \citep{pageHinkely1954}, Kolmogorv-Smirnov \citep{RAAB2020ReactiveSoftKS}, Hoeffding  bounds \citep{blanco2015HDDM} or PAC learning \citep{gama2004LearningDriftDetectiona,bifet2007LearningTimeChangingData}.  
Other methods such as \cite{yu2022MetaADDMetalearningBased} train a neural network to predict a concept drift and avoid hypothesis tests.
However, unlike \cara, these methods only detect concept drift and ignore the covariate drifts in both data and query streams.
Furthermore, unlike \cara, they do not balance the trade-off between the cost of retraining and make several unnecessary \retrain decisions due to false positives. 

\spara{Model selection methods}
These methods have multiple pre-existing trained models which are reused in the future.
They monitor the data stream and select the best model from the pre-existing set.
\citet{pesaranghader2016FrameworkClassificationData} use an error, memory and runtime (EMR) measure which they balance to select the best model.
\citet{mallick2022MatchmakerDataDrift} propose a method that can handle both covariate and concept drifts while picking the best model.
While such methods explore the trade-off between different costs, they do not consider the decision to retrain, but rather choose from a fixed set of pre-existing models.
By contrast, \cara addresses the problem of retraining \ml~models.

\spara{Data aware retraining}
\citet{zliobaite2015CostsensitiveAdaptationWhen} propose a {Return on Investment} (ROI) metric to decide when a \retrain decision is useful.
They show that when the gain in performance is larger than the resource cost a model should be updated.
However, unlike our work, they only perform an offline analysis and suggest monitoring the ROI during online evaluation. 
\citet{jelencic2022KLADWINEnhancedConcept} modify the approaches of drift detection with KL-divergence to make a combined metric and assess the optimal time to retrain a model.
Their results show that retraining when both approaches detect a drift results in better model performance.
However, unlike our work, they do not consider the resource costs of retraining the ML model while making decisions.

\section{Preliminaries}
\label{sec:problem-def}

% \subsection{Learning Paradigm}
In this section, we introduce the terms and notations that are necessary for the presentation of our technical contribution.

\spara{Data}
Each data entry $(\x,\y)$ consists of a point $\x\in\R^\feat$ and a target \y.
We use \X and \Y to refer to a set of points and their respective targets. 
In what follows, we assume a stream setting in which the data $\data=(\X,\Y)$ arrive over time in batches, with batch \ttime denoted by $\Dt=\round{\X_\ttime,\Y_\ttime}$.

\textit{Data drift} occurs when the arriving data change over time.
There are two main types of data drift as defined by \citet{gama2014SurveyConceptDrift}.
The first is \textbf{covariate drift} (or virtual drift), when only the distribution of the points $p(\x_t)$ changes with time.
This occurs when points in different batches come from different regions of the feature space. 
The second is \textbf{concept drift}, when the conditional distribution of the targets $p(\y_t|\x_t)$ changes with time.
This occurs when the underlying relationship between the target and the points is different across different data batches.
Furthermore, based on the frequency and duration of the change, concept drift can be categorized according to different patterns such as sudden, incremental, recurring, etc.

\spara{Queries}
An individual query is a point in a \feat-dimensional space, $\q\in\R^\feat$, without a ground-truth target.
In the streaming setting queries also arrive at every batch \ttime denoted by $\q_t\in\Qt$.
As queries do not have any ground-truth target \y, query streams may only exhibit a covariate drift when the query distribution $p(\q_\ttime)$ changes with time.

% \note[Ananth]{Will stick with Model as I use decision function to denote what is learned in the retraining algorithm}
\spara{Model}          
We use the term model to refer to a function $\model: \Xset \rightarrow \Yset$, where $\Xset\in\R^\feat$ and \Yset is the set of targets for the problem.
%
% The model captures the relationship between a point \x and the target \y.
%
Models are trained in a supervised manner using data \data that contain both points \x and ground-truth targets \y.
Once trained, a model is used to predict the target incoming query points $\hat{\y} := \model(\q)$. 
In the stream setting, $\Mt$ denotes a model trained using the data from batch $\ttime$ i.e., \Dt. 
% 
% Drifts in the data stream the affect the  

\spara{Retraining Algorithm}
% Why need for a retraining algorithm
% Due to data drift, the performance of an ML model varies with time. 
%
A retraining algorithm decides whe\-ther to retrain the ML model or not.
%  to optimize the trade-off between the performance costs of a stale model and the resource cost of retraining.
%
% Towards this, we define and quantify these two costs in \Cref{sec:costs}.
%
Formally, it consists of a decision function \decision with parameters \algoparams.
At each batch \ttime, the decision function \decision receives the data \Dt, queries \Qt and existing model \Mtprime (trained at batch $\tprime<\ttime$).
\decision makes either a \retrain or \keep decision based on its input and parameters \algoparams, where \algoparams typically consists of cost-related thresholds.
A \retrain decision indicates that the model must be retrained, while the \keep decision indicates that the existing model will be kept without retraining.
In what follows, we denote decisions of retraining algorithms as $\decision\round{\ttime,\tprime,\algoparams}\in\braces{\keep,\retrain}$.
%
% The retraining algorithms we discuss in this work aim to make decisions that optimize the resulting costs.
\begin{figure*}[ht]
    \centering
    \includegraphics[width=\textwidth]{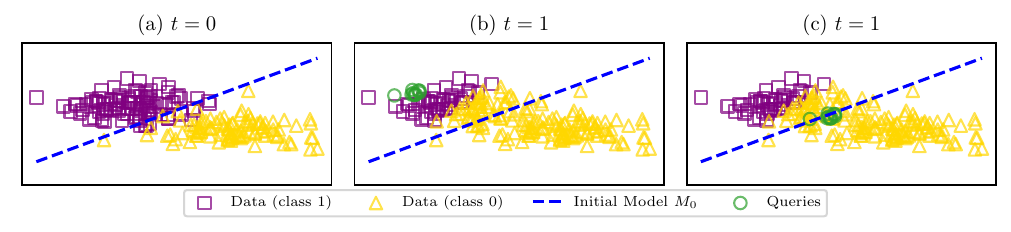}
    \caption{First scenario. (a) Initial data $\data_0$ and model $\model_0$. Concept drift occurs at $t=1$. (b) Queries are far from misclassifications. (c) Queries are close to misclassifications.  }
    \label{fig:example-scenarios}
\end{figure*}

\begin{figure*}[h]
    \centering
    \includegraphics[width=\textwidth]{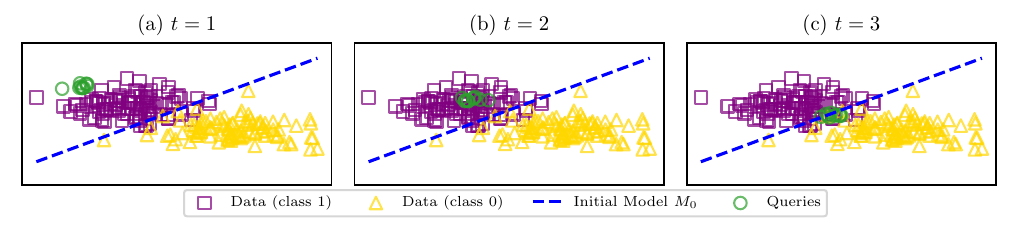}
    \caption{Second scenario. (a)-(c) Data has no concept or covariate drift in batches $t=1$ till $t=3$. Queries show covariate shift, moving from being far from the decision boundary in (a) to being closer to the misclassifications in (c).}
    \label{fig:example-scenarios-2}
\end{figure*}
\section{Problem Formulation}
\label{sec:costs}

In this section, we define the costs considered when making a \retrain or \keep decision and formalize the cost-optimization problem we address.
First, in Section~\ref{sec:stalenesscost}, we motivate and define the cost of model staleness in terms of expected query performance.
Second, in Section~\ref{sec:retrainingcost}, we define the model retraining cost which represents the resources required to retrain a model on a batch of new data. 
Third, in Section~\ref{sec:costmatrix}, we define a cost matrix that is computed offline retrospectively, given all the data and query batches within a given time interval, and which represents the possible costs for all the different possible retraining decisions in that time interval.
Based on this cost matrix, we define the cost of any retraining strategy, i.e., any sequence of \retrain \& \keep decisions.
Lastly, in Section~\ref{sec:problem-statement}, and based on the aforementioned definitions, we define the problem statement. 

% \begin{figure}[h]
%     \centering
%     \includegraphics[width=\columnwidth]{figs/poc/mini_proof_of_concept.pdf}
%     \caption{First example scenario. (a) Initial data $\data_0$ and model $\model_0$. Concept drift occurs at $t=1$. (b) Queries are far from misclassifications. (c) Queries are close to misclassifications.  }
%     \label{fig:example-scenarios}
% \end{figure}

\subsection{Staleness Cost}
\label{sec:stalenesscost}

The performance of a model potentially changes as the data undergoes a data drift.
In this paper, we aim to quantify the cost of keeping a model trained on old data, referred to as the ``staleness cost'', in terms of the model's query performance.
Let us `build' the formal definition of this cost as we discuss two exemplary scenarios that demonstrate how the performance of a model may change for different cases of data drift and queries.

In the first scenario, shown in  \Cref{fig:example-scenarios}, we consider a linear classification model trained on the initial 2D data at batch $\ttime=0$ in \Cref{fig:example-scenarios}(a).
Then, in the following batch $\ttime=1$, a concept drift occurs as seen in \Cref{fig:example-scenarios}(b) and (c) changing the distribution of class labels.
This leads to the stale model $\model_0$ misclassifying some new data points.
Typical concept drift methods will detect the drift in the data and make a \retrain decision ignoring the query distribution.
However, the performance of the model does depend on the query distribution of the queries on which it's called to make a prediction, and therefore so does the cost associated with a potential performance loss.
Specifically, if the query region is far from the misclassification region, as in \Cref{fig:example-scenarios}(b), then a \retrain decision will yield only small improvements in query performance, and so the staleness cost of a decision to \keep the current model is small in this case.
Conversely, if the query region is closer to the misclassification region, as in \Cref{fig:example-scenarios}(c), then a \retrain decision may improve query performance significantly, and so the staleness cost of a decision to \keep the current model is large in this case.

To distinguish between these two cases, we consider the model's misclassifications in the vicinity of the queries.
Towards this end, we define the model staleness cost for a single query point \q given a model \model and data \data as 
\begin{equation}
    \label{eq:qdm-single}
    \QDMq(\q,\data,\model) = \frac{1}{|\data|} \sum_{(\x,\y)\in\data} \simfn(q,x)\cdot\loss(\model,\x,\y), 
\end{equation} 
where $\simfn(\q,\x)$  is the similarity between the query and data point and $\loss(\model,\x,\y)$ is the loss of the model on the data point \x and labels \y.
The $\simfn(\cdot,\cdot)$ function captures the probability that a query's label is similar to a data point and is typically defined as a function of their distance.
Hence, the staleness cost as defined in \Cref{eq:qdm-single} measures the expected loss of a single query in the region of the data given a model.
%
% Note, as queries have no ground-truth labels, \Cref{eq:qdm-single} assumes that the data concept $p(\y|\x)$ and the query concept $p(\y|\q)$ are similar.
Back to the scenario of \Cref{fig:example-scenarios}, the staleness cost of the queries of \Cref{fig:example-scenarios}(b) would be small, because the misclassifications (large \loss) occur in areas away from the queries (small \simfn).
Conversely, the staleness cost of the queries of \Cref{fig:example-scenarios}(c) would be large, because the misclassifications (large \loss) occur in areas near the queries (large \simfn).

Building upon \Cref{eq:qdm-single}, we define the absolute staleness cost of all queries \Qt in batch \ttime given data \Dt and model \Mtprime trained at batch $\tprime\leq\ttime$ as the sum of all the individual \QDMq costs and is defined as 
\begin{equation}
    \label{eq:qdm-all}
    \QDM\round{\Qt,\Dt,\Mtprime} = \sum_{\q\in\Qt} \QDMq\round{\q,\Dt,\Mtprime}.
\end{equation}
More generally, we use $\QDM_{\ttime_1,\ttime_2,\ttime_3}$ to denote $\QDM\round{\query_{\ttime_1},\data_{\ttime_2},\model_{\ttime_3}}$ -- the absolute staleness cost of using model from batch ${\ttime_3}$ for the queries in batch $\ttime_1$ given the data from batch $\ttime_2$. 
%
% Typically  $\ttime_3\leq\min\round{\ttime_1,\ttime_2}$.

% \begin{figure}[h]
%     \centering
%     \includegraphics[width=\columnwidth]{figs/poc/mini_proof_of_concept2.pdf}
%     \caption{Second example scenario. Static data distribution in (a)-(c). Queries show covariate shift, moving from being far from the decision boundary in (a) to being closer to the misclassifications in (c).}
%     \label{fig:example-scenarios-2}
% \end{figure}

In the second scenario, shown in \Cref{fig:example-scenarios-2}, the data stream exhibits no drift from $\ttime=1$ to $\ttime=3$, but the query stream does drift over three batches.
Specifically, in each batch, the query distribution moves towards the decision boundary of the linear classifier $\model_0$.
Because of this, the absolute staleness cost \QDM, as defined in \Cref{eq:qdm-all}, increases from $\ttime=1$ to $\ttime=3$, because the number of data misclassifications increase as the queries move closer to the decision boundary.
However, \revision{if the data distribution is static}, retraining would not improve the query performance: as the data remain the same, the model that would be trained on them would also be the same.
To identify such a scenario and prevent unnecessary \retrain decisions, we measure the relative \emph{increase} in the staleness cost for the queries between two batches of data.
Towards this end, given the current batch \ttime and the batch when the model was trained $\tprime\leq\ttime$, we define a \emph{relative staleness cost} as  
\begin{equation}
    \label{eq:QDM-diff}
    \QDMdiff_{\ttime,\tprime} := \QDM_{\ttime,\ttime,\tprime} - \QDM_{\ttime,\tprime,\tprime}.
\end{equation}
Here, $\QDMdiff_{\ttime,\tprime}$ measures the relative increase in staleness cost of using model \Mtprime for the queries \Qt from \Dtprime (data used to train the model) to \Dt (data from current batch).

\subsection{Retraining Cost}
\label{sec:retrainingcost}
Resources such as time and monetary budget are spent when retraining a ML model.
The amount of resources spent depends on various factors such as the class of ML model, the dimensionality of the dataset, etc. 
We define retraining cost denoted by \revision{$\retraincost_\ttime$} as the relative value of the resources spent in the common units of the model staleness cost \revision{at batch \ttime}.
For example, assume a model takes 5sec to train, and the staleness cost measures the expected number of query misclassifications.
Then, \revision{$\retraincost_\ttime=150$} indicates that 5sec is equivalent to 150 query misclassifications i.e., one is willing to trade 1sec of time for 30 query misclassifications.
Therefore, larger values of \revision{$\retraincost_\ttime$} signifies resources are more important than model staleness cost and vice-versa. 
In the real world, the \revision{$\retraincost_\ttime$} is use-case dependent and is set by the administrator of the ML system. 
%
% For simplicity, we consider a static retraining cost, i.e., $\retraincost_t=\retraincost$ when computing a cost matrix.
% Therefore, the retraining cost provides cost-awareness to a retraining algorithm.

\subsection{Cost Matrix and Strategy Cost}
\label{sec:costmatrix}
Having defined the staleness and retraining costs, let us now collect in one matrix \costmatrix the costs associated with the possible decisions at different times $\ttime$ to \keep or \retrain a model that was trained at time \tprime.
Specifically, given a set of queries and data \Dt and \Qt where $0\leq\ttime\leq\T$ and models $\Mtprime$ where $\tprime\leq\ttime$ we define an upper-triangular cost matrix \costmatrix as follows 
\begin{singlespace}
    
    \begin{equation}
        \label{eq:cost-matrix}
        \Cost[\tprime,\ttime] =
        \begin{cases}
            \QDMdiff_{\ttime,\tprime} \quad& \text{if } \tprime< t\\
            \revision{\retraincost_\ttime} \quad& \text{if } \tprime=\ttime\\
            \infty \quad&  \text{otherwise} 
        \end{cases}    
        ,
    \end{equation}
\end{singlespace}
As defined, the upper-diagonal entries correspond to batches $\ttime$ for which a decision was made to \keep a model trained from a previous batch $\tprime<\ttime$, incurring a staleness cost $\QDMdiff_{\ttime,\tprime}$ but no retraining cost.
Moreover, the diagonal entries correspond to batches $\ttime=\tprime$ for which a decision was made to \retrain a model, incurring a retraining cost \revision{$\retraincost_\ttime$} but no staleness cost, as the new model is based on the latest data (i.e., $\QDMdiff_{t,t}=0$).
%

% \subsection{Strategy and Strategy Cost}
% \begin{algorithm}[tb]
%     \caption{Online Evaluation}
%     \label{alg:online-evaluation}
%  \begin{algorithmic}
%     \STATE {\bfseries Input:} Data \data, Queries \query, Decision Function \decision, parameters \algoparams
%     \STATE {\bfseries Output:} Retraining Strategy \strategy
%     \STATE Start online evaluation with $\tprime\leftarrow\T_{\offline}+1$ 
%     \STATE $\strategy \leftarrow \emptyset$
%     \STATE Train model \Mtprime with \Dtprime
%     \FOR{$\ttime\leftarrow\T_\offline+1$ {\bfseries to} $\T$}
%         \STATE $decision \leftarrow \decision(\ttime,\tprime,\algoparams)$
%         \IF{$decision = \retrain $}
%             \STATE Retrain model with \Dt and $\tprime \leftarrow \ttime$
%         \ENDIF 
%         \STATE $\strategy \leftarrow \strategy \cup \braces{\tprime}$
%     \ENDFOR
%     \STATE {\bf Return } \strategy
%  \end{algorithmic}
%  \end{algorithm}

 \begin{algorithm}[tb]
    \caption{Retraining Algorithm}
    \label{alg:retraining-algo-strategy}
 
    \revision{
    \begin{algorithmic}
    \STATE {\bfseries Input:} Data \data, Queries \query, Decision Function \decision, parameters \algoparams, \Tstart, \Tend
    \STATE {\bfseries Output:} Retraining Strategy \strategy
    \STATE Start evaluation with $\tprime\leftarrow\Tstart$ 
    \STATE $\strategy \leftarrow \emptyset$
    \STATE Train model \Mtprime with \Dtprime
    \FOR{$\ttime\leftarrow\Tstart$ {\bfseries to} $\Tend$}
        \STATE $decision \leftarrow \decision(\ttime,\tprime,\algoparams)$
        \IF{$decision = \retrain $}
            \STATE Retrain model with \Dt and $\tprime \leftarrow \ttime$
        \ENDIF 
        \STATE $\strategy \leftarrow \strategy \cup \braces{\tprime}$
    \ENDFOR
    \STATE {\bf Return } \strategy
 \end{algorithmic}
    }
 \end{algorithm}

A \emph{retraining strategy} is a sequence of the decisions made at each batch \revision{as seen in \Cref{alg:retraining-algo-strategy}}.
Formally, we define a strategy as the sequence of the model used at every batch $\ttime \in [0,\T]$
\begin{equation}
    \label{eq:strategy}
    \strategy = \round{s_0,s_1,\dots,s_\ttime,\dots,s_\T},
\end{equation}
where $s_\ttime\in[0,\ttime]$ denotes that the model $\model_{s_t}$ was used at batch $t$.
Hence, if $s_\ttime=\ttime$ then a \retrain decision was made at batch $\ttime$ otherwise a \keep decision was made.
The cost of a given retraining strategy is the sum of the cost for every batch defined as  
\begin{equation}
    \label{eq:strategy-cost}
    \cost{\strategy} = \sum_{\ttime=0}^{\T} C[s_\ttime,\ttime].
\end{equation}

\subsection{Problem Statement}
\label{sec:problem-statement}

For a retraining algorithm, we assume two phases, an \textbf{offline optimization phase} and an \textbf{online evaluation phase}.

In the offline phase historical streaming data and queries are available.
Typically, these data and queries would be collected and stored during a prior run of the system.
Let the data \Dt and queries \Qt corresponding to batches $0\leq\ttime\leq\Toffline$ be available in the offline phase.
During this phase, the complete offline cost matrix can be computed using \Cref{eq:cost-matrix} and analyzed.
The goal during this phase is to learn the patterns in the streams and optimize the parameters of the retraining algorithm. 

In the online phase batches $\Toffline < \ttime \leq \Tonline$ arrive sequentially.
\revision{The retraining algorithm is evaluated using \Cref{alg:retraining-algo-strategy} with $\Tstart=\Toffline+1$ and $\Tend=\Tonline$.}
At each batch $\ttime$ the decision function \decision is called with its, possibly optimized, parameters.
If a \retrain decision is made then a new model is trained otherwise the existing model is kept.
At the end of the evaluation, we obtain the online retraining strategy and its corresponding online strategy cost.

\begin{problem}
    Learn a decision function \decision in the offline phase which minimizes the online retraining strategy cost.   
\end{problem}
\section{\algo: Cost-Aware Retraining Algorithm}
% \label{sec:retraining-algorithms}
% A retraining algorithm consists of a decision function \decision and its corresponding parameters \algoparams.
% %
% At each batch \ttime, the function receives the data \Dt, queries \Qt and current model \Mtprime (trained at batch $\tprime<\ttime$).
% %
% Along with its parameters \algoparams the function chooses a  \retrain or \keep decision.
% %
% For simplicity, we use the notation as follows $\decision\round{\ttime,\tprime,\algoparams}\in\braces{\keep,\retrain}$.

% During the online phase the retraining algorithm are evaluated as seen in \Cref{alg:online-evaluation}.
% %
% The initial model is  trained on the data from batch  $\tprime=\Toffline$ and the evaluation begins from the following batch.
% %
% At each batch \ttime, the decision function \decision is called. 
% %
% Upon receiving a \retrain decision the model is retrained with 
% data from the current batch and \tprime is updated.
% %
% At the end of the evaluation, we obtain the retraining strategy.

% In this section, we present two retraining algorithms.
% %
% The first is the \oracle algorithm which produces the optimal retraining strategy retrospectively.
% %
% The second is our retraining algorithm \algo and its variants which are heuristics-based.
% %
% These variants aim to perform similar to the \oracle algorithm during online evaluation. 
\label{sec:our-algo}
\revision{
In this section, we present functions three cost-aware decision functions \decision, and their parameters \algoparams.
Each decision function gives rise to a variant of the retraining algorithm described in \Cref{alg:retraining-algo-strategy}.
We collectively refer to the three variants as Cost Aware Retraining Algorithm, \cara.
The variants are cost-aware because their parameters are chosen during the offline phase to optimize the trade-off between model staleness and retraining costs.
}

\revision{
During the offline phase each retraining algorithm finds parameters \algoparamsopt that minimize the offline strategy cost as follows:
\begin{equation}
    \label{eq:offline-optimisation}
    \algoparamsopt = \argmin_{\algoparams} \cost{\strategy},
\end{equation}
where \strategy is the retraining strategy from \Cref{alg:retraining-algo-strategy} with $\Tstart=0$ and $\Tend=\Toffline$.
This optimization is possible because all the data and queries in the offline phase are available in advance, as described in \Cref{sec:problem-statement}.
These optimal parameters \algoparamsopt balance the trade-off between the staleness and retraining costs and are then used with the decision function for the online evaluation.
In practice, we perform the optimization using techniques such as grid search over the space of all parameters. 

}

\subsection{\algoThresh: Threshold \revision{Variant}}
The \algoThresh variant has a threshold parameter \thresh and its decision function \decision is defined in \Cref{eq:threshold-heuristic}.
\begin{singlespace}
    
    \begin{equation}
        \label{eq:threshold-heuristic}
        {\small
        \decision(\ttime,\tprime,\braces{\thresh}) = \begin{cases}
            \keep  \quad& \QDMdiff_{\ttime,\tprime} < \thresh \\
            \retrain \quad& \text{otherwise}.
        \end{cases}
        }
    \end{equation}
\end{singlespace}
%
% During the offline phase, \algoThresh learns an optimal threshold parameter \threshopt by minimizing offline strategy cost.
%
% During the online phase, \decision is called with the current batch \ttime, the previous retraining batch \tprime and the optimal threshold $\algoparamsopt=\braces{\threshopt}$.
%
% Then, if the current cost, i.e., $\costmatrix\squares{\ttime,\tprime}$ is larger than the optimal threshold a \retrain signal is sent, otherwise a \keep signal is sent.
% In the offline phase an optimal threshold \threshopt is found for the given dataset and retraining cost.
%
During the online phase, if the current model staleness cost is larger than the threshold a \retrain decision is made.

\subsection{\algoCummThresh: Cumulative Threshold \revision{Variant}}
The \algoCummThresh variant \revision{inspired by \citep{page1954CUSUM}} has a cumulative threshold parameter \cumthresh and its decision function \decision is defined in \Cref{eq:cumulative-threshold-heuristic}.
\begin{singlespace}
    
    \begin{equation}
        \label{eq:cumulative-threshold-heuristic}
        {\small
        \decision(\ttime,\tprime,\braces{\cumthresh}) = \begin{cases}
            \keep \quad& \sum_{j=\tprime+1}^{\ttime}\QDMdiff_{j,\tprime} < \cumthresh \\
            \retrain \quad& \text{otherwise}
        \end{cases}
        }
    \end{equation}
\end{singlespace}

% During the offline phase the optimal cumulative threshold \cumthreshopt is found which minimizes the offline strategy cost.
%
% In the offline phase an optimal cumulative threshold \cumthreshopt is found for the given dataset and retraining cost.
%
During the online phase a variable tracks the cumulative cost, adding the current staleness cost at every batch from the last \retrain decision i.e., \tprime until the current time \ttime.
If this cumulative cost is larger than the parameter, then a \retrain decision is made and the cumulative cost is reset to $0$.

\subsection{\algoPeriod: Periodic \revision{Variant}}
The \algoPeriod variant has a periodicity \period and initial offset \offset.
During the online phase, the algorithm makes a \retrain decision every \period batch after an initial \offset number of batches.
The decision function is defined in \Cref{eq:periodic-heuristic}.   
\begin{singlespace}
    
    \begin{equation}
        \label{eq:periodic-heuristic}
        {\small
        \decision(\ttime,\tprime,\braces{\period,\offset}) = \begin{cases}
            \retrain \quad& (\ttime-\offset)\%\period =0\\
            \keep \quad& \text{otherwise}
            
        \end{cases}
        }
    \end{equation}
\end{singlespace}
\algoPeriod is a generalized version of a static periodic adaptation policy described in \citet{zliobaite2015CostsensitiveAdaptationWhen} also used in many real-world frameworks such as TensorFlow Extended \cite{tfx-retraining}. 
The difference is that \algoPeriod optimizes for the trade-off between the model staleness and retraining costs for a given dataset to find the optimal periodicity instead of using a pre-defined periodicity. 

\revision{
\subsection{Complexity Analysis}
Each \cara variant computes the model staleness cost \QDMdiff at each batch \ttime during online evaluation.
For analysis purposes, assume that the size of a batch of data and queries are \dataBatch and \queryBatch, respectively.
Then, the time complexity is $\bigObound(\queryBatch\dataBatch)$, dominated by the pairwise similarity computation, and the memory complexity is $\bigObound(\queryBatch+2\dataBatch)$ for storing the data and queries.
}

\section{The \oracle Algorithm}
\label{sec:oracle-alg}

In this section we present the \oracle algorithm which returns the optimal retraining strategy retrospectively.
The algorithm requires complete knowledge of all future batches of queries and data.
Given this knowledge, the algorithm finds the retraining strategy which has the lowest strategy cost.
In our experiments, we use the \oracle algorithm in the online phase to provide a lower bound for the strategy cost of any retraining algorithm.

Assume an oracle provides the optimal strategy as a set of the batches when a \retrain decision should occur, denoted by \oracleretrains.
Then the decision function follows the oracle as seen in \Cref{eq:oracle-decision-function}. 
\begin{singlespace}
    
    \begin{equation}
        \label{eq:oracle-decision-function}
        \decision\round{\ttime,\tprime,\braces{\oracleretrains}} =
        \begin{cases}
            \retrain \quad& \text{if } \ttime \in \oracleretrains \\
            \keep \quad& \text{otherwise}
        \end{cases}
    \end{equation}
\end{singlespace}
To find the optimal \retrain decision set \oracleretrains, we first define the optimal strategy as
\begin{equation}
    \label{eq:opt-strategy-cost}
    \strategy^* = \argmin_{\strategy}\cost{\strategy}.
\end{equation}
Then, we formulate a dynamic programming (DP) problem to find the optimal strategy cost and subsequently the optimal strategy.

Let \dpcost{\ttime,\prev} be the strategy cost at batch \ttime using a model trained at batch \prev. 
Then, the optimal strategy cost at \ttime is 
\begin{equation}
    \label{eq:dp-1}
    \dpcost[*]{t} = \min_{\prev\in[0,\ttime]} \dpcost{t,p},
\end{equation}
where the term \dpcost{\ttime,\prev} term is defined as
\begin{equation}
    \label{eq:dp-2}
    \dpcost{t,p} = \sum_{\tprime=\prev}^{t} \Cost[\prev,\tprime] + \dpcost[*]{p-1}.
\end{equation}  
The first term in \Cref{eq:dp-2} is the total cost of using the model trained at batch \prev in the batches from $p$ to $t$.
The second term in \Cref{eq:dp-2} is the optimal strategy cost prior to the last model retraining.
This sets up the recursive formulation for the DP problem.
Therefore, solving for $\dpcost[*]{\T}$ yields the optimal strategy cost at the end of batch \T.

\begin{algorithm}[tb]
    \caption{Memoize DP Table}
    \label{alg:dp}
 \begin{algorithmic}
    \STATE {\bfseries Input:} Cost Matrix \Cost, number of batches \T
    \STATE {\bfseries Output:} Memoized DP table \dptable
    \STATE Initialize \dptable as $(\T+1)\times(T+1)$ matrix filled with $\infty$
    \FOR{$\ttime=0$ {\bfseries to} $\T$}
    \STATE $\dptable[t,0] \leftarrow \sum_{\tprime=0}^{t} \Cost[0,\tprime]$ \COMMENT{fill first row}
    \ENDFOR
    \FOR{$\ttime\leftarrow 1$ {\bfseries to} $\T$}
    \FOR{$\prev\leftarrow 1$ {\bfseries to} $\ttime$}
        \IF{$\ttime = \prev$}
        \STATE $\dptable[t,\prev] \leftarrow C[t,t] + \min_{\tprime} \dptable[t-1,\tprime]$ 
        \ELSE
        \STATE $\dptable[t,\prev] \leftarrow C[\prev,t] + \dptable[t-1,\prev]$ 
        \ENDIF
    \ENDFOR
    \ENDFOR
    \STATE {\bfseries Return:} \dptable
 \end{algorithmic}
 \end{algorithm}

 \begin{algorithm}[tb]
    \caption{Oracle Retrains}
    \label{alg:opt-strat}
 \begin{algorithmic}
    \STATE {\bfseries Input:} DP table \dptable, number of batches \T
    \STATE {\bfseries Output:} Oracle \retrain decision batches \oracleretrains
    \STATE $\prev \leftarrow \argmin_{\prev^\prime} \dptable[T,\prev^\prime]$
    \STATE $\oracleretrains  \leftarrow \{\prev\}$
    \WHILE{$\prev>0$}
    \STATE $\prev \leftarrow \argmin_{\prev^\prime}\dptable[\prev-1,\prev^\prime]$
    \STATE $\oracleretrains  \leftarrow \oracleretrains \cup \{\prev\}$
    \ENDWHILE
    \STATE {\bf Return} \oracleretrains
 \end{algorithmic}
 \end{algorithm}

\Cref{alg:dp} describes a top-down method to memoize the strategy costs \dpcost{\cdot,\cdot} in the form of a DP table \dptable using the complete cost matrix \costmatrix.
Then, \Cref{alg:opt-strat} uses the computed \dptable and returns the oracle set \oracleretrains.
Note, \oracleretrains is a partial strategy consisting of only the \retrain decision batches. 
The optimal strategy $\strategy^{*}$ can be obtained by expanding \oracleretrains to include the \keep decisions. 

\revision{
% \subsection{Complexity Analysis}
The \oracle algorithm requires $\bigObound(\T^2)$ memory to materialize the memoized table \dptable and the cost matrix \Cost, where \T is the number of batches.
The algorithm has three stages, namely computing \Cost, memoizing \dptable and returning the oracle retrains \oracleretrains.
Therefore, the overall running time complexity is $\bigObound(\T^2\dataBatch\queryBatch + \T^2 + \T) = \bigObound(\T^2\dataBatch\queryBatch)$, where \dataBatch and \queryBatch are the size of a batch of data and query. 
}

\section{Experimental Setup}
\label{sec:setup}

We evaluate our \cara algorithm against the \oracle algorithm and other baselines on several synthetic and real-world datasets shown in \Cref{tab:datasets}.
In our experiments, we have fixed batch sizes and therefore consider a static retraining cost, i.e., $\retraincost_\ttime = \retraincost$.
We focus on the task of binary classification, i.e., $\Yset=\braces{0,1}$.
Hence, to compute the \QDMdiff cost, we use the radial basis function (RBF) kernel for similarity and the $0$-$1$ loss function defined as follows
\begin{align}
    \label{eq:sim-and-loss}
    \simfn(\q,\x) &= \exp\round{-\gamma\|\q-\x\|^2},\\
    \loss\round{\model,\x,\y} &= \mathcal{I}\round{\model(\x)\neq \y}
\end{align}
where $\gamma$ is the inverse of the variance and $\mathcal{I}$ is the indicator function.
%

% For each dataset, we analyze the performance of each variant of \algo described in \Cref{sec:our-algo}.
% %
% We compare the variants of \algo with the gold-standard \oracle algorithm and the baseline \neverretrain algorithms.

\begin{table}[ht]
    \centering
    \caption{Dataset statistics. $N$ is number of points, $d$ is number of dimensions, $\Toffline$ and $\Tonline$ indicate the offline and online batches retrospectively.}
    \label{tab:datasets}
    \begin{threeparttable}
    \begin{tabular}{lcccccc}
        \toprule
        Name &  $N$ & \feat &  $\T_{\offline}$ & $\Tonline$  \\
        \midrule
        \gauss & $100\,000$ & $2$ & $25$ & $100$  \\
        \Circle & $100\,000$& $2$ & $25$ & $100$ \\
        \covcon & $100\,000$ & $2$ & $25$ & $100$  \\
        \midrule
        \elec   &$45\,312$& $6$& $25$ & $100$ \\  
        \airlines  & $539\,383$ & $7198$\tnote{*} & $25$ & $100$ \\
        \covertype  & $581\,000$ & $54$ & $25$ & $100$ \\
        \bottomrule
    \end{tabular}
    \begin{tablenotes}
        \item[*] 5 nominal attributes into 7196 one-hot and 2 numeric features
    \end{tablenotes}
\end{threeparttable}
    
\end{table}

\subsection{Synthetic Datasets}
\label{sec:synthetic-datasets-setup}
We create several synthetic datasets with different concept and covariate drifts.
In each synthetic dataset, we generate $100$ batches and in each batch we create $1000$ data points and labels.
%
% See \Cref{app:synthetic-datasets} for more details.

\spara{\gauss} 
A 2D synthetic dataset that has recurring covariate shift.
Data points are sampled from a Gaussian whose centers move at each batch introducing covariate drift.
These points are labeled using a parabolic classification boundary.
At each batch \ttime, points $(\x_1,\x_2)$ are sampled from a Gaussian distribution with centers at $(c,0.5-c)$, where $c=((t+1)\%15)/30$.
The decision boundary for each point $\round{\x_1,\x_2}$ is fixed and given by $\x_2>4(\x_1-0.5)^2$.

\spara{\Circle \citep{pesaranghader2016FrameworkClassificationData}}
A 2D dataset that has a gradual concept drift.
Each data point has two features $(\x_1,\x_2)$ which are drawn from a uniform distribution. 
The classification is decided by the circle equation $\round{\x_1-c_1}^2+\round{\x_2-c_2}^2-r^2>0$, where $\round{c_1,c_2}$ is the circle center and $r$ is the radius.
The centers and radius of this circle are changed over time resulting in a gradual concept drift.

\spara{\covcon \citep{mallick2022MatchmakerDataDrift}}
A 2D dataset having both covariate and concept drift. 
In each batch \ttime, data points $(x_1,x_2)$ are drawn from a Gaussian distribution with mean $((t+1)\%7)/10$ and a fixed standard deviation of $0.1$ which introduces covariate drift.
The decision boundary of a data point is given by $\alpha*\sin(\pi\x_1)>\x_2$.
Every $10$ batches, the inequality of the decision boundary shifts from $>$ to $<$ and vice versa introducing a periodic concept drift.

\subsection{Real-World Datasets}
We experiment with three real-world datasets \elec, \covertype and \airlines which have unknown concept and covariate drifts.
For each dataset we split the data points into $100$ batches and use $25$ batches for the offline phase.

\spara{\elec \citep{Harries99splice-2comparative} } 
A binary classification dataset with the task to predict rise or fall of electricity prices in New South Wales, Australia. The data has concept drift due to seasonal changes in consumption patterns. 

\spara{\covertype \citep{covtype1999}}
Dataset containing $54$ cartographic variables of wilderness in the forests of Colorado and labels are the forest cover type.
We use binary version of the dataset from the LibSVM library \cite{LIBSVM}. 

\spara{\airlines  \citep{gomes2017AdaptiveRandomForests}}
Dataset with five nominal and 2 numerical features describing the airlines, flight number, duration, etc.\ of various flights.
The classification task is to predict if a particular flight will be delayed or not.
We use the version from the Sklearn Multiflow library \cite{Scikit-Multiflow}.
%
% See \Cref{app:real-world-datasets} for more details.

\subsection{Query Distributions}
\label{sec:query-distributions}
% \note[Ananth]{Do we require an additional non-static query distribution? }    
For each synthetic dataset, we generate $100$ queries in each batch \ttime from two different query streams.

In the first query stream, the queries are sampled from the data stream.
We randomly sample $10\%$ of the data entries in every batch and assign them as queries.
Here, the points $\x\in\X_t$ are used as queries \q in the experiment while the labels $\y\in\Y_t$ are used during evaluation.
We use the suffix ``\textsc{-D}'' to denote datasets with the first query stream for example, \covconData is the \covcon dataset with queries sampled from data stream.

In the second query stream, queries come from a static Gaussian distribution centered at $(0.5,0.5)$ with a standard deviation of $0.015$.
Here, we use the known concept to generate ground-truth labels \y for each query which are used during evaluation.
We use the suffix ``\textsc{-S}'' to denote datasets with the second query stream for e.g., \covconStatic is the \covcon dataset with static queries.
These different query streams will highlight the adaptability of the retraining algorithms to different query distributions.

For the real-world datasets, we only consider the first query stream where queries are sampled from the data stream.

\subsection{Baselines}
We compare the \algo variants against five baselines. 
%
% See \Cref{app:baselines} for more details.

\spara{\oracle} 
The retrospectively optimal retraining algorithm as presented in \Cref{sec:oracle-alg}.
We compute the online cost matrix and use the oracle algorithm to find the optimal retraining strategy.
The strategy cost of the \oracle baseline is a lower bound on the strategy cost for all retraining algorithms. 

\spara{\neverretrain} This is a trivial algorithm which always makes a \keep decision and never retrains the model.
Therefore, the decision function is $\decision(\ttime,\tprime) = \keep$.
The retraining strategy from this baseline is to always use the oldest trained model in a stream.
Moreover, as the retraining cost increases all other strategies should ideally devolve into the \neverretrain strategy.

\spara{\markov}
This retraining algorithm performs Markovian decisions based on only the current model staleness cost $\QDMdiff\round{\ttime,\tprime}$ and retraining cost \revision{$\retraincost_\ttime$}.
\begin{singlespace}
    
    \begin{equation}
        \label{eq:markov-heuristic}
        \decision(\ttime,\tprime) = \begin{cases}
            \keep \quad& \QDMdiff\round{\ttime,\tprime} < \revision{\retraincost_\ttime} \\
            \retrain \quad& \text{otherwise}
        \end{cases} 
    \end{equation}
\end{singlespace}
This baseline acts as an uncalibrated threshold algorithm where the threshold is always the model retraining cost \revision{$\retraincost_\ttime$}.

\spara{\adwin}
Proposed by \citet{bifet2007LearningTimeChangingData}, this detector maintains two sliding windows corresponding to old and new data.
When the windows differ beyond a specific threshold a drift is detected.
If a drift is detected in a batch then a \retrain decision is made. 

\spara{\ddm}
Proposed by \citet{gama2004LearningDriftDetectiona}, this detector uses the model's predictions of the data stream to monitor the online error rate.
With the assumption that classification rate will increase when the underlying concept changes, a drift is detected when the error rate increases beyond a certain threshold.
%
% When the error rate increases beyond a certain threshold a drift is detected.
%
If a drift is detected in a batch then a \retrain decision is made. 

\subsection{Evaluation Metrics}

For each retraining strategy we report the strategy cost and number of \retrain decisions along with two metrics. 

\spara{Query Accuracy} We perform a \emph{prequential} \cite{gama2014SurveyConceptDrift} i.e., test-then-train evaluation.
For example, if a \retrain decision is made in batch \ttime to update the model \Mtprime, the model is used first to answer queries \Qt and then retrained with data \Dt to produce the model \Mt.
In this manner we compute the query accuracy at every batch and then report the average query accuracy.
To achieve this, we stagger the strategy of a retraining algorithm by one batch to determine the model to be used in answering the queries.
Higher values are good and indicate that the retraining algorithm is able to answer queries effectively. 
% This is possible because we store the ground-truth labels for the query streams in both synthetic and real-world datasets as described in \Cref{sec:query-distributions}.

\spara{Strategy Cost Percentage Error (\scpe)}
We compute the relative difference between a retraining algorithm's strategy cost and the optimal strategy cost from the \oracle baseline. 
Percentage error is defined as $\text{PE} = 100\times{|(b-a)|}/{|b|}$, where $b$ and $a$ are the theoretical and experimental values respectively.  
Lower values are good and indicate that the retraining algorithm is closer to the optimal \oracle algorithm. 

\subsection{Implementation Details}
We use a Random Forest (RF) classifier from the Scikit-Learn library~\cite{scikit-learn} as the model.
Additionally, we present results using a Logistic Regression (LR) classifier in \Cref{app:LR-SGD-results}.
For each experiment we use five random seeds and average the results over the seeds.
The \adwin and \ddm baselines are implemented using Scikit-Multiflow~\cite{Scikit-Multiflow}.

For all \cara variants we select the length scale $\gamma$ of the RBF kernel based on the number of features \feat.
During the offline phase, we construct the offline cost matrix as described in \Cref{eq:cost-matrix}.
Then for the \algoThresh and \algoCummThresh variants, we use the dual annealing optimizer from SciPy~\cite{2020SciPy-NMeth} to find the optimal parameters \threshopt and \cumthreshopt.
For the \algoPeriod variant we perform a linear search over $\period\in [1,\Toffline]$ to find the optimal parameter \periodopt.
We use these optimal parameters to evaluate the \cara variant during the online phase.

All experiments were performed on a Linux machine with 32 cores and 50GB RAM.
Our code is available in the supplementary material.
\section{Experimental Results}
\label{sec:results}
\revision{In this section, we independently vary the staleness and retraining costs 
and analyze their trade-offs.}
%
% Due to space constraints, we only show results of \algo variants for the \scpe metric and results of \algoThresh, \oracle \& \adwin baselines for the average query accuracy metric.
% %
% Please see \Cref{app:results} for the full results.  

\begin{figure}[h]
    \centering
    
    \begin{subfigure}{0.45\textwidth}
        \centering
        \includegraphics[width=\textwidth,clip,trim={2mm 3mm 2mm 2mm}]{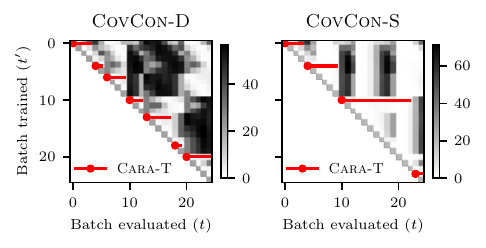}
        \caption{Offline cost matrices for CovCon datasets at  $\retraincost=20$}
        \label{fig:covcon-cost-matrices}
    \end{subfigure}
    \begin{subfigure}{0.45\textwidth}
        \centering
        \includegraphics[width=\columnwidth,clip,trim={2mm 3mm 2mm 2mm}]{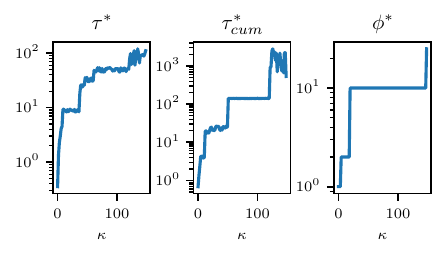}
        \caption{Optimal parameters for \algo variants for \covconData}
        \label{fig:covcon-optimal-params}
    \end{subfigure}
    \caption{(a) Varying query distribution. The red lines indicate the offline optimal \algoThresh strategy (markers and solid lines correspond to \retrain and \keep decisions respectively). (b) Varying the retraining cost. }
    % \label{<label>}
\end{figure}
\subsection{Varying Staleness Cost}
\label{sec:varying-staleness-analysis}
\revision{
We study the effect of the model staleness cost on the decisions of a retraining algorithm.
Towards this, for a given dataset, we keep the retraining cost \retraincost fixed and vary the query distribution to change the model staleness cost.
}
% We keep the retraining cost fixed at $\retraincost=20$ and change the staleness cost by varying the query distributions for the \covcon dataset.

% \begin{figure}[ht]
%     \centering
%     \includegraphics[width=0.5\columnwidth,clip,trim={2mm 3mm 2mm 2mm}]{figs/covcon_half_cost_matrices.pdf}
%     \caption{Offline cost matrices for \covcon datasets. Retraining cost was fixed at $\retraincost=20$. The red lines indicate the offline optimal \algoThresh strategy (markers and solid lines correspond to \retrain and \keep decisions respectively).}
%     \label{fig:covcon-cost-matrices}
% \end{figure}
%
\revision{
\Cref{fig:covcon-cost-matrices} presents the offline cost matrices at \revision{$\retraincost=20$} for the \covconData and \covconStatic datasets, which have non-static and static query distributions, respectively.
We also show the retraining strategies of the \algoThresh variant after finding the optimal threshold \threshopt from the cost matrices.
These retraining strategy are shown in the figure using a set of solid lines and markers which correspond to the \keep and \retrain decisions, respectively.}

There are three main takeaways.
First, the staleness cost captures both covariate and concept drift, seen by the gray regions and black checkerboard regions in the cost matrix of \covconData.
Second, the staleness cost is query-aware, i.e., cost increases only when the queries are affected by concept or covariate drift.
\revision{
We see this query-awareness in \covconStatic, where fewer regions have a high staleness cost due to the static query distribution.
Thirdly, for the given retraining cost, \algoThresh adapts to \covconStatic making far fewer \retrain decisions than in \covconData.    
}

\begin{figure*}[h]
    \centering
    \includegraphics[width=0.72\textwidth]{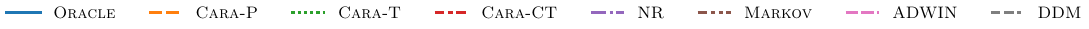}
    \includegraphics[width=0.72\textwidth,clip,trim={2mm 3mm 2mm 2mm}]{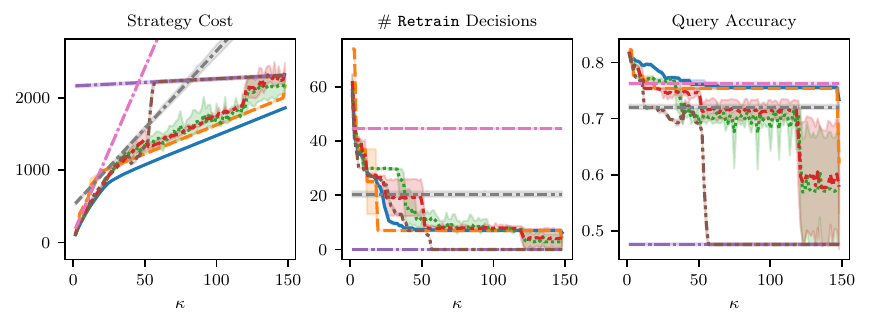}
    \caption{Strategy cost, number of retrains and query accuracy as a function of retraining cost for the \covconData dataset.}
    \label{fig:covcon-data-online}
\end{figure*}
\subsection{Varying Retraining Cost}
\revision{
We study the effect of the retraining cost on the retraining decision by fixing the query distribution and varying \retraincost for the \covconData dataset.

First, for a range of \retraincost, we find the optimal parameters \algoparamsopt for each \cara variant using the offline cost matrix for every retraining cost as seen in \Cref{fig:covcon-optimal-params}.
These optimal parameters implicitly control the decisions the retraining algorithm makes in the online phase.
For example, when $\retraincost\leq 50$, \algoPeriod makes a \retrain decision every other batch because the optimal periodicity $\periodopt=2$.
However, when $50\leq\retraincost\leq 125$, fewer \retrain decisions will be made because $\periodopt=10$. 
}
%
% Furthermore, as described in \Cref{sec:synthetic-datasets-setup}, the \covconData dataset has a synthetic concept drift every ten batches.
% %
% This indicates that during offline optimization \algoPeriod learns the exact periodicity of the data distribution for a certain range of retraining cost. 

% \begin{figure}[h]
%     \centering
%     \includegraphics[width=0.5\columnwidth,clip,trim={2mm 3mm 2mm 2mm}]{figs/covcon_data_optimized_parameters.pdf}
%     \caption{Optimal parameters for \algo variants as a function of retraining cost for the \covconData  dataset.}
%     \label{fig:covcon-optimal-params}
% \end{figure}

Next, we evaluate every retraining algorithm in the online phase using the optimal parameters for a range of \retraincost (see \Cref{app:online-evaluation}). 
In \Cref{fig:covcon-data-online}, we present the evaluation metrics for the \covconData dataset as a function of the retraining cost \retraincost.
There are three main takeaways.
First, \cara variants have strategy costs close to the optimal \oracle baseline, \revision{effectively balancing the cost-trade-off.}
Second, as the retraining cost increases, \cara implicitly reduces the number of \retrain decisions by minimizing for strategy cost.
Third, the query accuracy of \cara is comparable to \adwin and \ddm baselines, except in regions of very high retraining cost, where it has high variance. 

\begin{figure*}[h]
    \centering
    \includegraphics[width=0.72\textwidth]{figs/legend.pdf}
    \begin{subfigure}{0.72\textwidth}
        \centering
        \includegraphics[width=\textwidth,clip,trim={2mm 3mm 2mm 2mm}]{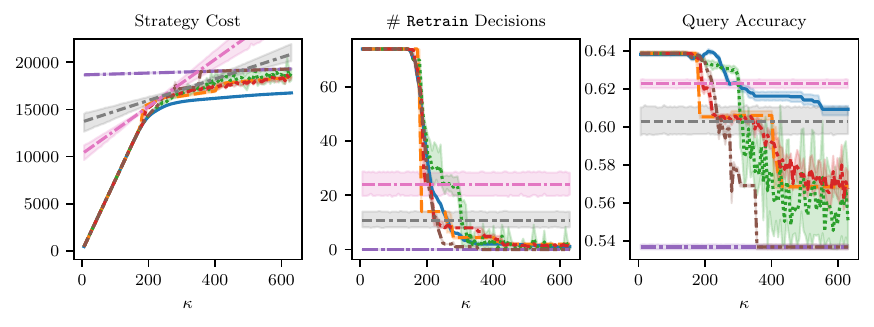}
        \caption{\airlines}
        \label{fig:airlines-results}
    \end{subfigure}
    \begin{subfigure}{0.72\textwidth}
        \centering
        \includegraphics[width=\textwidth,clip,trim={2mm 3mm 2mm 2mm}]{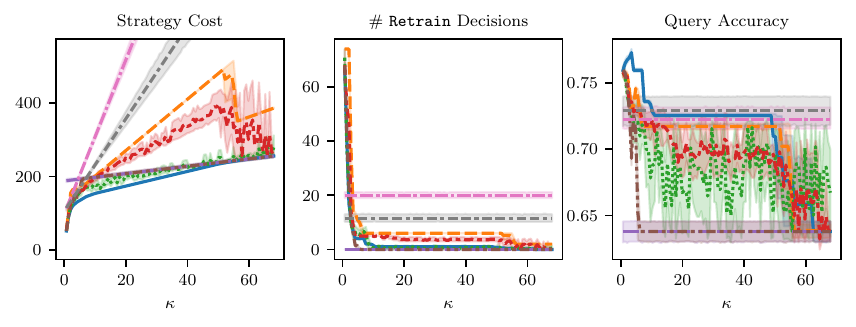}
        \caption{\elec}
        \label{fig:elec-results}
    \end{subfigure}
    \begin{subfigure}{0.72\textwidth}
        \centering
        \includegraphics[width=\textwidth,clip,trim={2mm 3mm 2mm 2mm}]{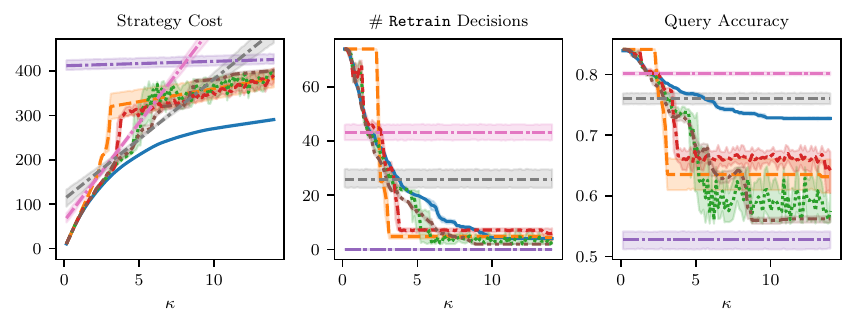}
        \caption{\covertype}
        \label{fig:covertype-results}
    \end{subfigure}
    \caption{Strategy cost, number of retrains and query accuracy as a function of retraining cost for the real-world datasets.}
    \label{fig:real-world-results}
\end{figure*}
\subsection{Real-World Data Analysis}
\revision{
We analyze the online performance of the \cara variants in the three real-world datasets for a range of retraining costs \retraincost.
\Cref{fig:real-world-results} presents detailed online evaluation metrics as a function of \retraincost for the \airlines, \elec and \covertype datasets.}
There are three main observations.

\revision{
First, \cara variants learn to retrain more frequently in regions of low retraining cost, similar to the \oracle baseline.
This retraining results in higher query accuracy compared to the drift detection baselines.
Second, based on the dataset, \cara variants adapt to retrain less frequently in regions of moderate retraining cost.
For example, when $\retraincost>2.5$, all \cara variants (and the \oracle baseline) choose to retrain fewer than ten times in the \elec dataset.
Whereas, in the \covertype dataset, \cara variants drop to fewer than ten retrains only when the retraining cost is relatively higher.
Third, \cara variants are very conservative in regions of high retraining cost and retrain infrequently.
Here, we also observe a gap between the \oracle baseline and \cara variants in strategy cost and query accuracy while the number of \retrain decisions remains similar.
This performance gap indicates \cara variants make few suboptimal \retrain decisions that in high-cost regions.
Hence, there is scope for improvement through designing more complex variants of the \cara algorithm.}

\section{\revision{Discussion}}
\label{sec:discussion}
\revision{
This section presents the aggregated results for all retraining algorithms and datasets and discusses the key takeaways.
First, \Cref{sec:strategy-cost-discussion} compares each retraining algorithm against the \oracle baseline and reports the average percentage error in strategy cost.
Second, \Cref{sec:query-acc-num-retrain-discussion} compares the average query accuracy against the number of \retrain decisions taken for each algorithm.
Lastly, \Cref{sec:cara-variants-discussion} compares and discusses the different \cara variants' performance.
}
\begin{table*}[h]
    \centering
    \small
    \caption{ Average \scpe for all algorithms and datasets. Lower is better.}
    \label{tab:mean-scpe-all}
    \begin{tabular}{lrrrrrrr}
\toprule
Dataset &  \textsc{Cara-T} &  \textsc{Cara-CT} &  \textsc{Cara-P} &  \textsc{NR} &  \textsc{Markov} &  \textsc{ADWIN} &  \textsc{DDM} \\
\midrule
\textsc{CovCon-D}    &            17.72 &             19.26 &             \textbf{16.3} &       141.12 &            40.85 &          163.05 &         71.81 \\
\textsc{CovCon-S}    &            \textbf{15.58} &             22.94 &           104.79 &       191.38 &            25.17 &          305.52 &        106.03 \\ \midrule
\textsc{Circle-D}    &            \textbf{45.61} &             47.29 &            84.96 &       350.75 &           185.16 &          108.59 &        115.83 \\
\textsc{Circle-S}    &            \textbf{33.89} &             45.15 &            79.34 &       305.75 &           193.38 &          132.77 &        109.38 \\ \midrule
\textsc{Gauss-D}     &             \textbf{8.63} &             59.93 &           135.95 &        17.05 &             9.67 &          468.27 &        306.49 \\
\textsc{Gauss-S}     &            \textbf{66.62} &             96.38 &           202.09 &       169.52 &           105.51 &          628.08 &        433.44 \\ \midrule \midrule
\textsc{Electricity} &             \textbf{8.32} &             40.32 &            64.07 &        17.08 &            11.01 &          285.78 &        156.55 \\
\textsc{Airlines}    &             \textbf{6.11} &              5.48 &             5.45 &       141.13 &             8.99 &           86.41 &         99.95 \\
\textsc{Covertype}   &            26.39 &             27.92 &            38.37 &       188.85 &            \textbf{25.61} &           73.63 &         57.12 \\
\bottomrule
\end{tabular}

\end{table*}

\subsection{Strategy Cost Results}
\label{sec:strategy-cost-discussion}
We present the mean \scpe over the retraining costs in \Cref{tab:mean-scpe-all}.
There are three main takeaways from these results.
First, \cara variants have the lowest \scpe amongst all algorithms in almost all datasets.
\revision{Even in the \covertype dataset, where the \markov baseline has the lowest \scpe, the \algoThresh variant is close to the \markov baseline.}
Second, \algoThresh has the least \scpe amongst all \cara variants consistently.
In the \covconData and \airlines datasets, \algoPeriod has a lower \scpe, \revision{indicating an intrinsic periodicity} in the underlying data or query distribution.
However, in other datasets that do not present periodicity, such as \covconStatic and \gaussStatic, the \algoPeriod algorithm performs worse than other \cara variants due to unnecessary \retrain decisions.
Third, the \adwin and \ddm baselines always have a high mean \scpe due to their inability to adapt to the change in query distribution and retraining costs.

\begin{table*}[h]
    \centering
    \small
        \caption{Average query accuracy for all algorithms and datasets. Higher is better.}
        \label{tab:mean-query-acc-all}
        \begin{tabular}{lrrrrrrrr}
\toprule
Dataset &  \textsc{Cara-T} &  \textsc{Cara-CT} &  \textsc{Cara-P} &  \textsc{Oracle} &  \textsc{NR} &  \textsc{Markov} &  \textsc{ADWIN} &  \textsc{DDM} \\
\midrule
\textsc{CovCon-D}    &              0.7 &              0.71 &             \textbf{0.76} &             \textbf{0.76} &         0.48 &             0.56 &            \textbf{0.76} &          0.72 \\
\textsc{CovCon-S}    &             0.79 &              0.79 &             0.61 &              0.8 &         0.47 &             0.66 &            \textbf{0.85} &          0.81 \\ \midrule
\textsc{Circle-D}    &             0.96 &              \textbf{0.97} &             0.95 &             \textbf{0.97} &         0.85 &             0.89 &            \textbf{0.97} &          0.94 \\
\textsc{Circle-S}    &             0.93 &              0.93 &             0.91 &             \textbf{0.96} &         0.91 &             0.92 &            0.95 &          0.92 \\ \midrule
\textsc{Gauss-D}     &              0.9 &              0.88 &              0.8 &             \textbf{0.92} &          0.9 &              0.9 &            0.88 &           0.9 \\
\textsc{Gauss-S}     &             0.99 &              0.99 &             0.98 &              \textbf{1.0 }&          \textbf{1.0 }&              \textbf{1.0 }&            0.93 &          0.85 \\ \midrule \midrule
\textsc{Electricity} &             0.68 &              0.69 &              0.7 &             0.71 &         0.64 &             0.64 &            0.72 &          \textbf{0.73} \\
\textsc{Airlines}    &              0.6 &               0.6 &              0.6 &             \textbf{0.62} &         0.54 &             0.58 &            \textbf{0.62} &           0.6 \\
\textsc{Covertype}   &             0.66 &               0.7 &             0.67 &             0.76 &         0.53 &             0.66 &             \textbf{0.8} &          0.76 \\
\bottomrule
\end{tabular}

\end{table*}
\begin{table*}[h]
    \centering
    \small
        \caption{Average number of \retrain decisions for all algorithms and datasets.}
        \label{tab:mean-num-retrains-all}
        \begin{tabular}{lrrrrrrrr}
\toprule
Dataset &  \textsc{Cara-T} &  \textsc{Cara-CT} &  \textsc{Cara-P} &  \textsc{Oracle} &  \textsc{NR} &  \textsc{Markov} &  \textsc{ADWIN} &  \textsc{DDM} \\
\midrule
\textsc{CovCon-D}    &            13.89 &             13.33 &            10.75 &             11.2 &            0 &             7.18 &            44.8 &          20.4 \\
\textsc{CovCon-S}    &            10.05 &             10.22 &             9.68 &             8.21 &            0 &             5.99 &            44.8 &          20.4 \\ \midrule
\textsc{Circle-D}    &             5.67 &              7.49 &             5.41 &             3.89 &            0 &             2.13 &            11.0 &           4.2 \\
\textsc{Circle-S}    &             3.56 &              6.06 &             5.41 &             3.09 &            0 &             1.42 &            11.0 &           4.2 \\ \midrule
\textsc{Gauss-D}     &             2.62 &              6.29 &             7.29 &             2.23 &            0 &             1.88 &            31.8 &          22.2 \\
\textsc{Gauss-S}     &             2.57 &              5.04 &             5.85 &             1.92 &            0 &             0.92 &            31.8 &          22.2 \\ \midrule \midrule
\textsc{Electricity} &             2.39 &              5.08 &             7.83 &             2.36 &            0 &             1.53 &            20.0 &          11.6 \\
\textsc{Airlines}    &            27.69 &             25.29 &            24.43 &            25.23 &            0 &            23.12 &            24.0 &          10.6 \\
\textsc{Covertype}   &            16.62 &              17.9 &            17.06 &            19.42 &            0 &            15.66 &            43.2 &          26.0 \\
\bottomrule
\end{tabular}

\end{table*}
\subsection{Query Accuracy \&  Number of \retrain Decisions}
\label{sec:query-acc-num-retrain-discussion}
\Cref{tab:mean-query-acc-all} and \Cref{tab:mean-num-retrains-all} present the mean query accuracy and number of \retrain decisions, respectively.
There are three takeaways.
First, the \oracle baseline has better or similar query accuracy than the drift detection baselines \adwin and \ddm, which retrain more frequently.
Second, the \cara variants have query accuracy close to the \oracle baseline. 
Lastly, looking at the average number of \retrain decisions, we see that \cara variants are very similar and only slightly higher compared to the \oracle baseline.
Furthermore, in datasets where the \neverretrain baseline has high query accuracy, such as \circlesStatic and \gaussStatic, we see that the \cara variants learn to make fewer \retrain decisions compared to the \adwin and \ddm baselines.

\subsection{Discussion on different  \cara variants}
\label{sec:cara-variants-discussion}
Each variant of \cara exhibits specific properties valuable in different use cases.

First, as defined in \Cref{eq:threshold-heuristic}, \algoThresh relies on a single threshold and, therefore, is highly sensitive to changes in the model staleness cost.
The variance of the optimal threshold \threshopt found during the offline phase (seen in \Cref{fig:covcon-optimal-params}), especially in regions of high retraining cost, is reflected in the variance of the metrics such as strategy cost and query accuracy (seen in \Cref{fig:covcon-data-online}).
Nevertheless, due to this sensitivity, \algoThresh has the lowest strategy cost amongst all the \cara variants across datasets and query distributions, as discussed in \Cref{sec:strategy-cost-discussion}. 

Second, \algoCummThresh makes a \retrain decision when the cumulative staleness cost crosses a threshold, as defined in \Cref{eq:cumulative-threshold-heuristic}.
Therefore, \algoCummThresh is conservative by design, making fewer \retrain decisions than \algoThresh while adapting to changing model staleness costs.
We see this conservative nature in the low variance of strategy cost and query accuracy in \Cref{fig:covcon-data-online,fig:real-world-results}.
However, this performance stability comes with a slightly higher strategy cost, as seen in \Cref{tab:mean-scpe-all}.

Third, \algoPeriod is a standard periodic adaptation policy.
As seen in \Cref{sec:varying-staleness-analysis}, \algoPeriod performs well and has a low strategy cost when the data (or queries) follow a periodic pattern.
However, when the data (or query) distribution does not exhibit an apparent periodicity, \algoPeriod makes several unnecessary \retrain decisions with a higher strategy cost.

% Based on these observations, a potential end-user of the proposed \cara algorithm performs an offline analysis with the different variants of \cara.
% 
% Then, they select the variant which shows the best performance for their given data and query distributions to be used in the online phase.
\section{Conclusion}
\label{sec:conclusion}
In our paper we studied the trade-off between the model staleness and retraining costs.
%
% We explore algorithms that balance this trade-off and produce a retraining strategy. 
% %   
% Towards this, we develop an \oracle algorithm which produces the optimal retraining strategy in retrospect.
%
Towards this, we motivated and defined the model staleness cost as the increase in the model's misclassifications in the region of the query.
Further, we measured the retraining cost in the same metric which allowed us to compute the strategy cost of a retraining algorithm.
We presented \algo, our cost-aware retraining algorithm which optimizes for the trade-off by minimizing the strategy cost.
Through our analysis on synthetic data we demonstrated that \algo variants are able to adapt to both concept and covariate drifts in data and query streams.
Furthermore, we show that \cara variants have lower number of \retrain decisions as a consequence of minimizing for strategy cost while having query accuracy comparable to drift detection baselines such as \adwin and \ddm. 
We also developed a retrospective optimal \oracle algorithm which we used as a baseline in our experiments as the lower bound the strategy cost of any retraining algorithm.
We observed that amongst all \cara variants, \algoThresh had the lowest strategy cost percentage error with respect to the \oracle baseline across both real-world and synthetic datasets. 
Lastly, in real-world datasets \algoThresh had query accuracy comparable to the \ddm and \adwin baselines while implicitly making fewer \retrain decisions.

Our current work has several limitations and potentials for future research.
First, the current implementation of the staleness cost computes pairwise similarities which has a considerable computational cost when the number of data points in batch are large.
Therefore, exploring the use of k-dimensional (KD) trees or core-sets will increase efficiency and improve scalability to larger datasets.  
Second, in regions of high retraining cost there is a gap between the performance of \cara variants and the \oracle baseline. 
Hence, using Reinforcement Learning (RL) based approaches to learn more complex decision functions during the offline optimization phase will lead to more robust retraining algorithms with better performance.
Lastly, in our experiments the retraining cost \retraincost was static, and the \cara variants were optimized in the offline phase for a given retraining cost.  
A future research direction is to study and develop retraining algorithms for a non-static retraining cost i.e., retraining cost which can change during each batch.

% \begin{enumerate}
%     \item Contributions
%     \begin{itemize}
%         \item Framework with metrics for both query and retraining cost awareness
%         \item Flexible extensible framework
%         \item Optimal \oracle algorithm
%         \item Variants of \algo that can be run online
%     \end{itemize}
%     \item Discussion and Future Work
%     \begin{itemize}
%         \item Make QDM computation efficient using KD trees and quantization
%         \item Develop more complex decision functions
%         \item Extend to an online adaptive strategy  
%     \end{itemize}
% \end{enumerate}

\section*{Acknowledgements}
Ananth Mahadevan would like to thank Arpit Merchant and Sachith Pai for their useful suggestions regarding the \oracle baseline.
Michael Mathioudakis is supported by University of Helsinki and Academy of Finland Projects MLDB (322046) and HPC-HD (347747).
%% If you have bibdatabase file and want bibtex to generate the
%% bibitems, please use
\bibliographystyle{elsarticle-num-names}
\bibliography{references}

%% The Appendices part is started with the command \appendix;
%% appendix sections are then done as normal sections
\appendix
\section{Methodology}

\subsection{Scenario 1}

\begin{figure*}[h]
    \centering
    {\includegraphics[width=0.8\textwidth,clip,trim={0 2.5mm 0 5.5mm}]{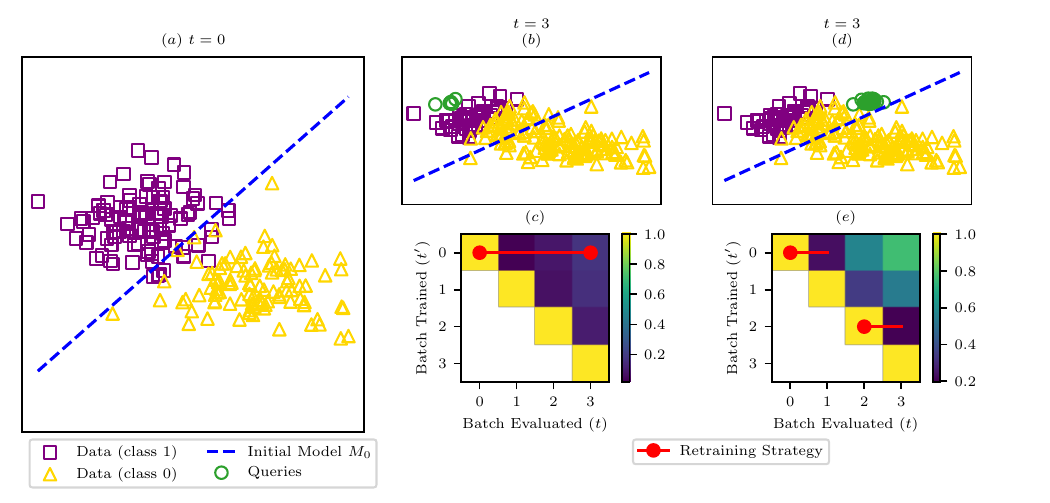}}
    \caption{Extended example scenario from \Cref{fig:example-scenarios}. (a) Initial data and model $\model_0$. (b) Queries are far from misclassification. (d) Queries are closer to misclassifications. (c) and (e): Cost matrix and \oracle retraining strategies with retraining cost fixed to $\retraincost=1$ for (b) and (d) respectively.}
    \label{fig:scenario-1-extended}
\end{figure*}

In \Cref{fig:scenario-1-extended}, we present an extended version of the scenarios discussed in \Cref{sec:costs}.
Here, the scenario here spans across four batches, the initial data at batch $\ttime=0$ shown in \Cref{fig:example-scenarios}(a) and the final batch at $\ttime=3$ shown in \Cref{fig:example-scenarios}(b) and (d).
At $t=3$, we see the data has clearly drifted away with more class 0 data points being on the incorrect side of the decision boundary of initial model $\model_0$.
In \Cref{fig:example-scenarios}(b), the queries are far away from the decision boundary and would therefore be classified correctly by $\model_0$.
We see in \Cref{fig:example-scenarios}(c) where the staleness cost for the first row corresponding to model $\model_0$ is low.
The \oracle retraining strategy indicated by the red line also suggests that the initial model need not be updated in this scenario.
In \Cref{fig:example-scenarios}(d), the queries are much closer to the decision boundary of $\model_0$ and the data misclassifications.
Therefore, we see in \Cref{fig:example-scenarios}(e) the staleness cost increases steadily in the first row.
The \oracle retraining strategy is to retrain the model at $t=2$ to reduce the staleness cost of keeping $\model_0$ at batches $t=2$ and $t=3$.
\begin{figure*}[h]
    \centering
    {\includegraphics[width=0.8\textwidth,clip,trim={0 2.5mm 0 2.5mm}]{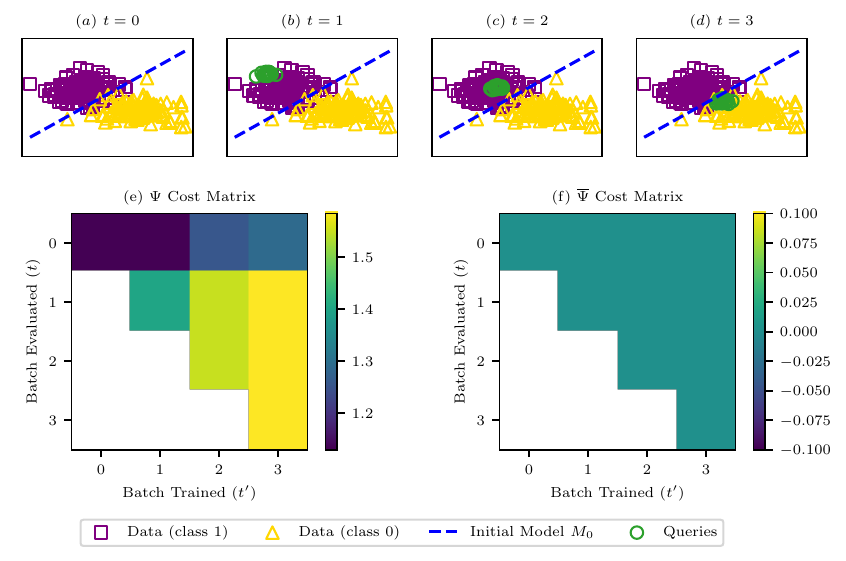}}
    \caption{Extended scenario 2 from \Cref{fig:example-scenarios-2}.(a)-(d) different batches of static data with moving query distribution. (e) Cost matrix computed using simple staleness \QDM (\Cref{eq:qdm-all}). (f) Cost matrix computed using staleness cost \QDMdiff (\Cref{eq:QDM-diff}). }
    \label{fig:scenario-2}  
\end{figure*}
\subsection{Scenario 2}
In \Cref{fig:scenario-2}, we provide an example of the second scenario discussed in \Cref{sec:costs}.
In \Cref{fig:scenario-2}(a)-(d) we see the query distribution moving towards the classification boundary of the initial model $\model_0$.
As expected, the simple staleness cost \QDM defined in \Cref{eq:qdm-single} increases with incoming batches.
However, in this scenario making a \retrain decision will not improve query performance as the data points and labels are static.
The staleness cost \QDMdiff defined in \Cref{eq:QDM-diff} mitigates this flaw.
We see that the \QDMdiff cost in \Cref{fig:scenario-2} is $0$ indicating there is no cost to keeping the initial model $\model_0$.

% \section{Exploring the Range of Retraining Costs}
% \label{app:very-retraining-costs}
% %
% The range of retraining costs analyzed for each dataset is determined from the offline cost matrix.
% %
% The minimum is always $0$.
% %
% The maximum is the retraining cost at which the \oracle strategy is the same as the \neverretrain strategy.  
% %
% This is identified by a simple line search using the \texttt{fsolve} optimizer from the scipy library.
% %
% This retraining cost range is the same when analyzing the online performance of a retraining algorithm. 

\begin{figure}[h]
    \centering
    \includegraphics[width=0.4\columnwidth]{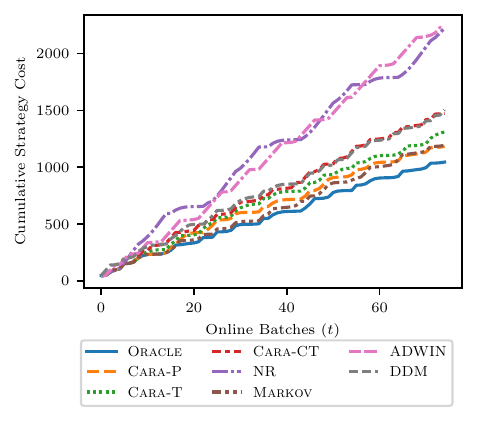}
    \caption{Online evaluation of all retraining algorithms for the \covconData dataset with $\retraincost=\fixedretraincost$. The plot shows the cumulative strategy cost as a function of the online batches. }
    \label{fig:covcon-data-cum-cost}
\end{figure}
\section{Online Evaluation}
\label{app:online-evaluation}
\revision{
We evaluate every retraining algorithm in the online phase using the optimal parameters for a range of \retraincost. 
In \Cref{fig:covcon-data-cum-cost}, we present the online evaluation of all retraining algorithms for the \covconData dataset at $\retraincost=46$, where the cumulative strategy cost is on the y-axis and the online batches are on the x-axis.
There are three takeaways.
First, the \neverretrain algorithm has the highest cost as it does not retrain, accumulating staleness cost.
Conversely, the \oracle algorithm has the lowest cost as it is optimal.
Second, the \algoPeriod variant is the closest to the \oracle strategy cost, performing similarly to the \markov baseline.
Third, although the \adwin makes several \retrain decisions, it has a strategy cost similar to the \neverretrain baseline.
This is because the \adwin baseline is not cost-aware and accumulates retraining costs from making excessive \retrain decisions.
}
%
% Lastly, we see for this value of \retraincost, that the \algo-CT and \ddm strategies perform similarly. 

\revision{ 
\section{Results with Logistic Regression}
\label{app:LR-SGD-results}
This section studies the impact of the choice of model class on the experiments and discussions from \Cref{sec:results,sec:discussion}. 
Towards this, we change the model \model from a Random Forest classifier to a Logistic Regression (LR) classifier.
We use the \texttt{SGDClassifier} from scikit-learn \cite{scikit-learn}  with loss parameter set to ``log''.
This loss parameter results in a LR classification model. 
Note, we do not perform an extensive grid-search and used the standard hyperparameters for the model.
Therefore, the resulting trained LR model might perform poorly for different data and query distributions. 
We share these result in \Cref{tab:LR-SGD-mean-scpe-all,tab:LR_SGD-mean-query-acc-all,tab:LR-SGD-mean-num-retrains-all}.
\begin{table*}[h]
    \centering
    \caption{ Average \scpe for all algorithms and datasets using a Logistic Regression model. Lower is better.}
    \label{tab:LR-SGD-mean-scpe-all}
    \begin{tabular}{lrrrrrrr}
\toprule
Dataset &  \textsc{Cara-T} &  \textsc{Cara-CT} &  \textsc{Cara-P} &  \textsc{NR} &  \textsc{Markov} &  \textsc{ADWIN} &  \textsc{DDM} \\
                     &                  &                   &                  &              &                  &                 &               \\
\midrule
\textsc{CovCon-D}    &            27.03 &             16.46 &            13.12 &       158.27 &            43.36 &          210.16 &         62.41 \\
\textsc{CovCon-S}    &            55.61 &            126.68 &           449.28 &       256.81 &            50.05 &          852.72 &        230.38 \\
\textsc{Circle-D}    &           840.88 &           1067.78 &          1170.57 &      2370.12 &          2306.45 &         1879.82 &       1576.84 \\
\textsc{Circle-S}    &          1635.49 &            626.12 &           1577.3 &      1664.42 &          1645.01 &         2379.48 &       1833.82 \\
\textsc{Gauss-D}     &           148.57 &             216.2 &           211.52 &       101.55 &            82.58 &          850.56 &        548.71 \\
\textsc{Gauss-S}     &           387.83 &            275.75 &          1398.26 &       564.41 &            559.1 &         4490.27 &       3006.65 \\
\textsc{Electricity} &           1031.5 &           1830.55 &          6310.27 &     14948.71 &           625.42 &         5230.54 &       3174.36 \\
\textsc{Airlines}    &            11.68 &             35.89 &            31.38 &        25.11 &            10.38 &          211.51 &         90.32 \\
\textsc{Covertype}   &            28.65 &             35.69 &            40.93 &       129.37 &            45.52 &          189.32 &         75.35 \\
\bottomrule
\end{tabular}

\end{table*}
\begin{table*}[h]
    \centering
        \caption{Average query accuracy for all algorithms and datasets using a Logistic Regression model. Higher is better.}
        \label{tab:LR_SGD-mean-query-acc-all}
        \begin{tabular}{lrrrrrrrr}
\toprule
Dataset &  \textsc{Cara-T} &  \textsc{Cara-CT} &  \textsc{Cara-P} &  \textsc{Oracle} &  \textsc{NR} &  \textsc{Markov} &  \textsc{ADWIN} &  \textsc{DDM} \\
                     &                  &                   &                  &                  &              &                  &                 &               \\
\midrule
\textsc{CovCon-D}    &             0.66 &              0.72 &             0.76 &             0.77 &         0.47 &             0.56 &            0.74 &          0.71 \\
\textsc{CovCon-S}    &             0.67 &              0.56 &             0.45 &             0.62 &         0.47 &             0.54 &            0.91 &           0.9 \\
\textsc{Circle-D}    &             0.81 &               0.8 &             0.81 &             0.77 &         0.82 &             0.82 &            0.81 &          0.82 \\
\textsc{Circle-S}    &             0.09 &              0.09 &             0.09 &             0.09 &         0.09 &             0.09 &            0.09 &          0.09 \\
\textsc{Gauss-D}     &             0.85 &              0.85 &             0.76 &              0.9 &         0.78 &             0.79 &            0.84 &          0.83 \\
\textsc{Gauss-S}     &              1.0 &               1.0 &             0.95 &              1.0 &          1.0 &              1.0 &             0.7 &          0.59 \\
\textsc{Electricity} &             0.67 &              0.67 &             0.64 &              0.7 &         0.56 &             0.67 &            0.65 &          0.67 \\
\textsc{Airlines}    &             0.56 &              0.57 &             0.56 &             0.57 &         0.55 &             0.56 &            0.59 &          0.56 \\
\textsc{Covertype}   &             0.63 &              0.66 &             0.66 &             0.72 &         0.47 &             0.54 &            0.75 &           0.7 \\
\bottomrule
\end{tabular}

\end{table*}
\begin{table*}[h]
    \centering
    \caption{Average number of \retrain decisions for all algorithms and datasets using a Logistic Regression model.}
    \label{tab:LR-SGD-mean-num-retrains-all}
    \begin{tabular}{lrrrrrrrr}
\toprule
Dataset &  \textsc{Cara-T} &  \textsc{Cara-CT} &  \textsc{Cara-P} &  \textsc{Oracle} &  \textsc{NR} &  \textsc{Markov} &  \textsc{ADWIN} &  \textsc{DDM} \\
                     &                  &                   &                  &                  &              &                  &                 &               \\
\midrule
\textsc{CovCon-D}    &            11.71 &             11.85 &             10.2 &            10.55 &            0 &             6.02 &            46.0 &          17.0 \\
\textsc{CovCon-S}    &             7.82 &              8.29 &             6.15 &             4.37 &            0 &              2.5 &            46.0 &          17.0 \\
\textsc{Circle-D}    &             2.32 &              2.22 &             4.24 &             3.23 &            0 &             0.57 &             6.8 &           5.0 \\
\textsc{Circle-S}    &             0.56 &              2.06 &             4.15 &             2.56 &            0 &             0.23 &             6.8 &           5.0 \\
\textsc{Gauss-D}     &             4.72 &              7.78 &             5.28 &             1.32 &            0 &             0.57 &            28.2 &          19.0 \\
\textsc{Gauss-S}     &             0.85 &              1.89 &             4.61 &             1.16 &            0 &             0.05 &            28.2 &          19.0 \\
\textsc{Electricity} &            18.48 &             14.13 &            10.56 &            13.57 &            0 &            20.46 &            13.2 &          10.0 \\
\textsc{Airlines}    &             6.52 &             10.06 &            10.13 &             5.51 &            0 &             4.72 &            29.4 &          11.6 \\
\textsc{Covertype}   &             12.2 &             12.59 &            11.11 &             9.29 &            0 &             5.77 &            43.6 &          17.0 \\
\bottomrule
\end{tabular}

\end{table*}

Comparing the \scpe in \Cref{tab:LR-SGD-mean-scpe-all} to \Cref{tab:mean-scpe-all}, we see that strategy cost errors for all retraining algorithms are higher when using LR classifier.
Furthermore, comparing \Cref{tab:LR_SGD-mean-query-acc-all} to \Cref{tab:mean-query-acc-all},the query accuracies are also lower for all algorithms.
For example, the query accuracy for the \circlesStatic dataset, is $0.09$ across algorithms indicating the LR model trained on the data was unable to answer the static queries accurately. 
These results indicate that the LR classifier is unable to effectively model the non-linearity in the data, especially for \circlesStatic and \elec datasets. 
Next, \cara variants make far fewer \retrain decisions compared to \adwin and \ddm baselines, as seen in \Cref{tab:LR-SGD-mean-num-retrains-all}.
Nevertheless, \cara variants have good query accuracy and low strategy cost.
Finally, the overall trends discussed in \Cref{sec:discussion} still hold, i.e., \algoThresh and \algoCummThresh have low strategy cost and their query accuracy is comparable to the \oracle baseline.
}

% \section{Results}
% \label{app:results}

\begin{figure*}[h]
    \centering
    \includegraphics[width=0.72\textwidth]{figs/legend.pdf}
    \centering
    \includegraphics[width=0.72\textwidth,clip,trim={2mm 3mm 2mm 2mm}]{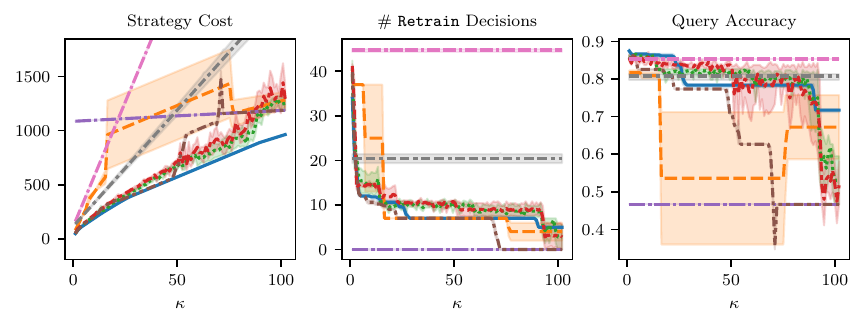}
    \caption{Strategy cost, number of retrains and query accuracy a function of retraining cost for the \covconStatic dataset.}
    \label{fig:covcon-online}
\end{figure*}

\begin{figure*}[h]
    \centering
    \includegraphics[width=0.72\textwidth]{figs/legend.pdf}
    \begin{subfigure}{0.72\textwidth}
        \centering
        \includegraphics[width=\textwidth,clip,trim={2mm 3mm 2mm 2mm}]{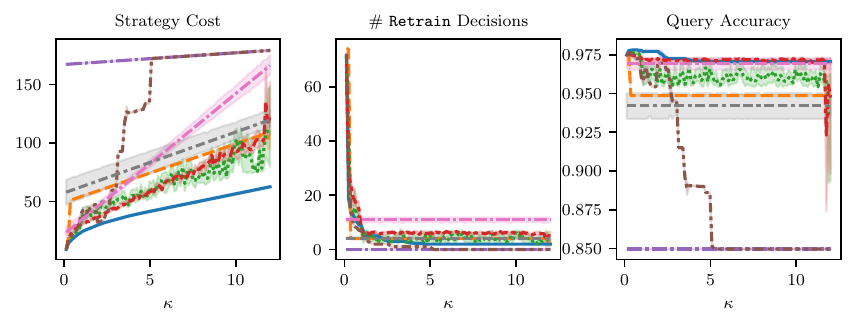}
        \caption{\circlesData}
        \label{fig:circles-data-results}
    \end{subfigure}
    \begin{subfigure}{0.72\textwidth}
        \centering
        \includegraphics[width=\textwidth,clip,trim={2mm 3mm 2mm 2mm}]{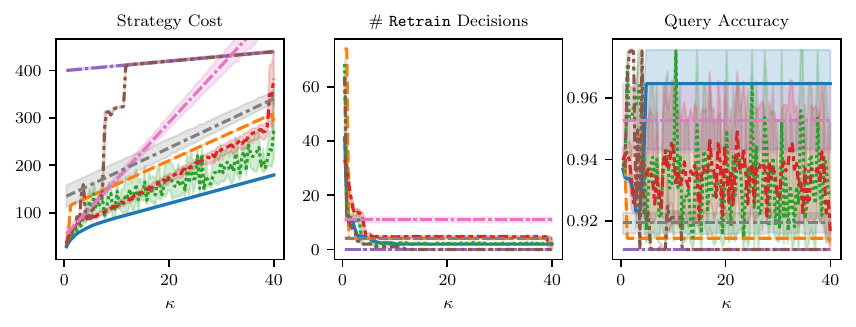}
        \caption{\circlesStatic}
        \label{fig:circles-static-results}
    \end{subfigure}
    \caption{Strategy cost, number of retrains and query accuracy as a function of retraining cost for the \circles dataset.}
    \label{fig:circles-online}
\end{figure*}

\begin{figure*}[h]
    \centering
    \includegraphics[width=0.72\textwidth]{figs/legend.pdf}
    \begin{subfigure}{0.72\textwidth}
        \centering
        \includegraphics[width=\textwidth,clip,trim={2mm 3mm 2mm 2mm}]{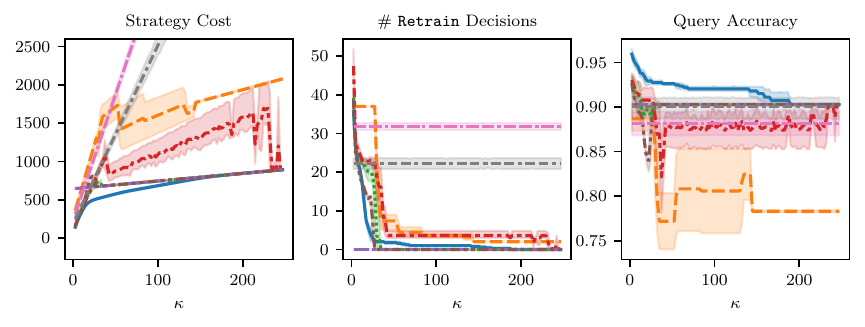}
        \caption{\gaussData}
        \label{fig:gauss-data-results}
    \end{subfigure}
    \begin{subfigure}{0.72\textwidth}
        \centering
        \includegraphics[width=\textwidth,clip,trim={2mm 3mm 2mm 2mm}]{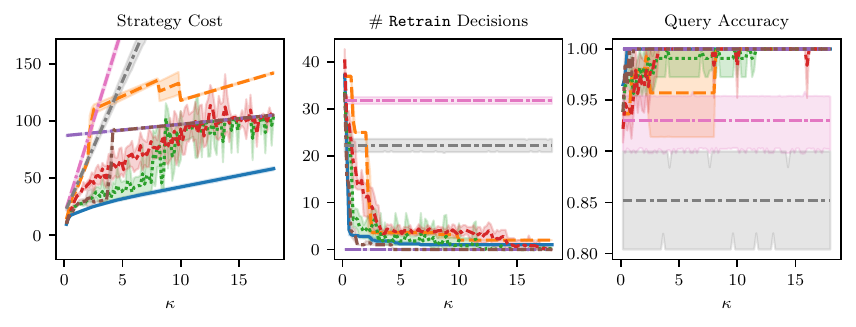}
        \caption{\gaussStatic}
        \label{fig:gauss-static-results}
    \end{subfigure}
    \caption{Strategy cost, number of retrains and query accuracy as a function of retraining cost for the \gauss dataset.}
    \label{fig:gauss-online}
\end{figure*}

% Here we present the complete results for all datasets, algorithms and baselines described in \Cref{sec:setup}.
\revision{
\section{Synthetic Dataset Results}
In this section, we present the results of all the synthetic datasets described in \Cref{tab:datasets}.

\Cref{fig:covcon-online} shows the results for the \covconStatic dataset which has a static query distribution.
Here, we see the ineffectiveness of the \algoPeriod variant show in the significantly larger strategy cost. 
This large strategy cost is due to unnecessary \retrain decisions being made even when the query distribution is static.
These decisions lead to accumulated retraining costs which add to the total cost of the \algoPeriod retraining strategy. 
Furthermore, the \retrain decisons are also poorly timed as we see the query accuracy of \algoPeriod is lower than other \cara variants.
From these results, we conclude that periodic retraining strategies cannot adapt well to aperiodic data and query changes.  

In \Cref{fig:circles-online}, we present the results for the \circles dataset.
We observe that the number of \retrain decisions required is lower compared to the \covcon dataset. 
Furthermore, the variance of the \algoThresh variant is seen in the query accuracy of \circlesStatic dataset, shown in \Cref{fig:circles-static-results}.
This variance is due to the sensitivity of the \algoThresh variant discussed in \Cref{sec:discussion}.

In \Cref{fig:gauss-online}, we present the results for the \gauss dataset.
We again see that \algoPeriod performs poorly in both \gaussData and \gaussStatic, as seen in \Cref{fig:gauss-data-results,fig:gauss-static-results}.
Interestingly, in \Cref{fig:gauss-static-results}, we observe that \adwin and \ddm baselines make frequent \retrain decisions, however, have slightly lower query accuracy than other retraining algorithms.
On the other hand, the \oracle baseline and the \cara variants make much fewer \retrain decisions, resulting in lower strategy cost, and have better query accuracy. 
Therefore, from these results, we conclude that the \oracle and \cara variants can adapt better to drift in data and query distributions while optimizing the trade-off between retraining costs and model staleness.    
}

% \subsection{Real-World Dataset Results}

% The detailed plots of all the metrics for the real-world datasets are presented in \Cref{fig:real-world-results}.

% \begin{figure*}
%     \centering
%     \includegraphics[width=\textwidth]{figs/legend.pdf}
%     \begin{subfigure}{\textwidth}
%         \centering
%         \includegraphics[width=\textwidth,clip,trim={0 3mm 0 2mm}]{figs/airlines_data_online.pdf}
%         \caption{\airlines}
%         \label{fig:airlines-results}
%     \end{subfigure}
%     \begin{subfigure}{\textwidth}
%         \centering
%         \includegraphics[width=\textwidth,clip,trim={0 3mm 0 2mm}]{figs/elec_data_online.pdf}
%         \caption{\elec}
%         \label{fig:elec-results}
%     \end{subfigure}
%     \begin{subfigure}{\textwidth}
%         \centering
%         \includegraphics[width=\textwidth,clip,trim={0 3mm 0 2mm}]{figs/covertype_data_online.pdf}
%         \caption{\covertype}
%         \label{fig:covertype-results}
%     \end{subfigure}
%     \caption{Strategy cost, number of retrains and query accuracy of different retraining algorithms as a function of retraining cost for the real-world datasets.}
%     \label{fig:real-world-results}
% \end{figure*}
%% else use the following coding to input the bibitems directly in the
%% TeX file.

% \begin{thebibliography}{00}

% %% \bibitem{label}
% %% Text of bibliographic item

% \bibitem{}

% \end{thebibliography}
\end{document}